\documentclass[11pt]{article}

\usepackage[preprint]{acl}

\usepackage{times}
\usepackage{latexsym}
\usepackage{float}
\usepackage[T1]{fontenc}

\usepackage[utf8]{inputenc}
\usepackage{amsmath}
\usepackage{amssymb}
\usepackage{orcidlink}
\usepackage{graphicx}
\usepackage{xcolor}
\usepackage{soul}

\usepackage{booktabs}
\usepackage{graphicx}
\usepackage{tabularx}
\usepackage{adjustbox}
\usepackage{float}
\usepackage{placeins}
\usepackage{inconsolata}
\usepackage{enumitem}

\title{Representation-Aware Unlearning via Activation Signatures: From Suppression to Entity-Signature Erasure}

\author{%
\parbox{\textwidth}{\centering
\textbf{Syed Naveed Mahmood}\textsuperscript{1},
\textbf{Md.~Rezaur Rahman Bhuiyan}\textsuperscript{1,*},
\textbf{Tasfia Zaman}\textsuperscript{1,*}\\
\textbf{Jareen Tasneem Khondaker}\textsuperscript{1},
\textbf{Md.~Sameer Sakib}\textsuperscript{1},
\textbf{K.~M.~Shadman Wadith}\textsuperscript{1}\\
\textbf{Nazia Tasnim}\textsuperscript{2},
\textbf{Farig Sadeque}\textsuperscript{1}\\[2pt]
\textsuperscript{1}Computer Science and Engineering, BRAC University, Dhaka, Bangladesh\\
\textsuperscript{2}Boston University, Boston, MA, USA\\
\textsuperscript{*}Equal contribution.
}}

\begin{document}
\maketitle

\begin{abstract}

Entity-level unlearning is usually evaluated by what a model says: whether it stops naming the target, refuses a query, or shifts a Truth Ratio distribution. These output-level tests, however, do not show whether a subject's internal representation has been attenuated. We introduce the \textbf{Entity Representation Unlearning Framework (ERUF)}, a representation-aware framework that mines subject-specific activation signatures, suppresses the corresponding activation direction, and distills the behavior into LoRA parameters. Among evaluated baselines, ERUF is the only method that jointly achieves surface level suppression, internal attenuation, and utility preservation. On TOFU \texttt{forget10}, ERUF achieves FQ $=0.99$ and MU $=0.62$, matching reported oracle utility while approaching oracle forget quality. Across most standard foundation-model settings, ERUF maintains low leakage and low internal target activation (SMR $=0.00$--$1.10\%$, EL10 $<0.06$, utility drift $<3\%$). On Llama-3.1-8B, adversarial entity recovery falls from 63.89\% to 20.15\%, while name-agnostic recovery decreases by 72.7--77.4\%. Joint surface/internal diagnostics further reveal scale-dependent behavior in reasoning-prior models that surface metrics alone would miss. We interpret these results as operational evidence of representation-level attenuation, not as a formal guarantee of irreversible deletion. Code is available in the anonymous repository\footnote{\href{https://anonymous.4open.science/r/Representation-Aware-Unlearning-via-Activation-Signatures-KIF-963D}{Anonymous repository}}.
\end{abstract}

\section{Introduction}
Large language models (LLMs) are increasingly deployed in real-world NLP systems, but their training corpora can contain sensitive, copyrighted, or otherwise removable information, creating tension with the GDPR right to erasure~\citep{gdpr2016, zhang2024rightforgotten}. LLMs store training information in distributed parameters \citep{10.1145/3744746}, making targeted removal technically difficult. \emph{LLM unlearning} aims to remove selected information while preserving general utility~\citep{yao-etal-2024-machine, ren2025sokmachineunlearninglarge}.
Entity-level unlearning can be especially difficult because the forget target is not a single answer string or isolated fact. A named subject can be recovered through associated facts, aliases, descriptions, relations, and contextual clues \citep{ma2025unveilingentitylevelunlearninglarge, jin2024rwku}. Thus, a model may suppress direct mentions while preserving an internal trace of the target that resurfaces through indirect access routes. Prior work warns that such apparent forgetting can reflect obfuscation rather than latent trace removal~\citep{sun2025unlearningobfuscation, ren2025sokmachineunlearninglarge}. This motivates treating entity-level unlearning as a representation-level task as much as an output-level one.


\begin{figure}[t]
  \includegraphics[width=\columnwidth]{images/motivation_image_eruf1.pdf}
  \caption{\textbf{Comparison of approaches.} Output-level unlearning can suppress the answer while leaving a latent subject trace. ERUF targets a mined subject signature and distills the suppression into a LoRA adapter.}
  \label{fig:motivation}
\end{figure}

We propose the \textbf{Entity Representation Unlearning Framework (ERUF)}, a representation-aware framework for entity-level unlearning. ERUF mines subject-linked activation signatures from real-world entity prompts, suppresses the corresponding activation direction through Signature Suppression Capsules, and distills the resulting behavior into a global LoRA adapter~\citep{hu2021loralowrankadaptationlarge}. The target behavioral profile is \textbf{surface-level suppression, internal attenuation, and utility preservation}. 

We evaluate this profile using Subject Mention Rate (SMR) for surface leakage, and EL10 for early subject-associated token probability mass relative to the base model. We treat EL10 as a proxy for latent target activation rather than proof of deletion, and corroborate it with adversarial, name-agnostic, hidden-state, and sequential diagnostics. Among evaluated baselines, no method simultaneously achieves low surface leakage, reduced EL10, and low utility drift, whereas ERUF achieves this three-way profile on most standard foundation models while the diagnostic explicitly identifies where it does not.
The behavioral difference is clearest under probes that omit or obscure the canonical subject name. On Llama-3.1-8B, ERUF reduces adversarial recovery from 63.89\% to 20.15\%, and reduces name-agnostic alias, keyword, and target-mass metrics by 72.7--77.4\%. These results, together with hidden-state, sequential, and cross-family analyses, support ERUF as operational evidence of representation-level attenuation rather than merely surface refusal.

Our contributions are:
\begin{enumerate}[topsep=1pt, itemsep=0pt, partopsep=0pt, parsep=0pt, leftmargin=*]
    \item A real-world entity prompt dataset for mining subject-linked activation signatures;
    \item ERUF, a representation-aware framework combining capsule-based suppression with LoRA distillation for entity-level unlearning;
    \item A dual-metric surface/internal evaluation protocol (SMR\,+\,EL10) distinguishing output suppression from latent target attenuation;
    \item An empirical behavioral analysis across evaluated baselines, model families, adversarial and name-agnostic probes, hidden-state diagnostics, and sequential unlearning.
\end{enumerate}

\section{Related Work}

\noindent\textbf{Benchmark-driven LLM unlearning.}
Standard LLM unlearning avoids full retraining by optimizing approximate forgetting objectives while preserving retained behavior, typically on benchmarked forget/retain tasks such as TOFU~\citep{maini2024tofu} and MUSE~\citep{shi2024musemachineunlearningsixway}. Optimization variants including NPO~\citep{zhang2024negativepreferenceoptimizationcatastrophic}, SimNPO~\citep{fan2024simplicity}, AltPO~\citep{mekala-etal-2025-alternate}, ReLearn~\citep{xu2025relearnunlearninglearninglarge}, and UnDIAL~\citep{dong-etal-2025-undial} are largely compared through benchmark behavior rather than direct tests of subject-linked internal activation. Recent audits argue that LLM unlearning evaluation should be more robust and multifaceted, since benchmark success can overstate practical forgetting under query variations and does not by itself show whether information has been removed from model weights~\citep{lynch2024methodsevaluaterobustunlearning,thaker2025positionllmunlearningbenchmarks,deeb2025unlearningmethodsremoveinformation}.

\noindent\textbf{Entity-level unlearning.}
Entity-level work studies removal of broader subject-level knowledge rather than isolated instances. Ma et al.~\citep{ma2025unveilingentitylevelunlearninglarge} formalize entity-level unlearning and show that forget-set coverage strongly affects success. RWKU~\citep{jin2024rwku} benchmarks real-world entities with adversarial and locality probes, while OPT-OUT~\citep{choi-etal-2025-opt} studies real-world entity removal with explicit retention of neighboring and world knowledge. ERUF shares this real-world entity setting, but focuses on whether suppressing subject outputs is accompanied by attenuation of subject-associated internal activation.

\noindent\textbf{Representation-level intervention and auditing.}
Knowledge localization and representation-geometry work suggests that factual associations are mediated by identifiable MLP/FFN computations and directions in representation space~\citep{meng2023locatingeditingfactualassociations,geva2021transformerfeedforwardlayerskeyvalue,park2024linearrepresentationhypothesisgeometry}. This motivates representation-level unlearning methods such as RMU~\citep{li2024wmdpbenchmarkmeasuringreducing}, LUNAR~\citep{shen2026llm}, and ReGLU~\citep{xiao2026representation}. Recent work also distinguishes unlearning from obfuscation and asks whether information has been removed from accessible weights or latent representations, rather than merely suppressed at the output surface~\citep{sun2025unlearningobfuscation,deeb2025unlearningmethodsremoveinformation}. ERUF follows this line by pairing SMR and EL10 with adversarial and name-agnostic probes, operationalizing the distinction between surface suppression and internal attenuation for entity-level targets.

\section{Methodology}
We introduce a three-stage pipeline: \textbf{(1)} localizing target knowledge through statistical analysis on extracted activation-signature , \textbf{(2)} suppressing the localized signature through lightweight gating modules we term \textbf{Signature Suppression Capsules}, and \textbf{(3)} distilling the capsule behavior into a global LoRA adapter \citep{hu2021loralowrankadaptationlarge}. The training objective combines preference-style supervision with token-level unlikelihood and stability regularization \citep{rafailov2024directpreferenceoptimizationlanguage,welleck2019neuraltextgenerationunlikelihood,Kirkpatrick_2017}. Figure~\ref{fig:pipeline} summarizes the pipeline. All implementation details are provided in Appendix \ref{sec:implementation}.

\begin{figure*}[t]
  \includegraphics[width=1\linewidth]{images/pipeline_diagram_clean_v4.pdf}
  \caption{\textbf{Entity Representation Unlearning Framework (ERUF) pipeline.} Entity prompts from the Real-World Entity Dataset (\ref{dataset}) probe MLP activations to mine subject-specific signatures via contrastive analysis. These signatures instantiate Signature Suppression Capsules (Section \ref{sec:cap_forge}) at high-salience inference layers. The UPU Loop (Section \ref{sec:loop}) then distills the resulting suppressed behavior into a global LoRA adapter for durable representation-level knowledge removal.}
  \label{fig:pipeline}
\end{figure*}

\subsection{Dataset Construction}
\label{dataset}
The localization stage requires activation profiling over controlled subject prompts. We construct a Real-World Entity Dataset grounded in factual triples $(\textit{subject}, \textit{predicate}, \textit{object})$ extracted from Wikipedia and Wikidata. Following the knowledge-tuple framing of \citet{meng2023locatingeditingfactualassociations}, we instantiate triples into templates such as ``Is it true that \texttt{\{subject\}}'s \texttt{\{predicate\}} was \texttt{\{object\}}?'' The dataset contains five probe types:
\begin{enumerate}[topsep=1pt, itemsep=0pt, partopsep=0pt, parsep=0pt, leftmargin=*]
    \item \textbf{Direct:} asks for a specific fact.
    \item \textbf{Contextual:} embeds the target fact in a broader information request.
    \item \textbf{Implicit:} uses confirmation-style prompts involving the target fact.
    \item \textbf{Reasoning:} asks the model to explain or reason about the fact.
    \item \textbf{Misleading:} pairs the subject with incorrect information to probe whether the model verifies or accepts a suggested falsehood.
\end{enumerate}
Full dataset statistics, prompt templates, and subject distributions are provided in Appendix~\ref{sec:dataset}.

\subsection{Localization via Activation-Signature Extraction}
\label{sec:activation}

Prior causal tracing work suggests that factual associations can be localized in MLP modules \citep{meng2023locatingeditingfactualassociations}. We therefore probe a 4-bit quantized model on the Real-World Entity Dataset and capture intermediate MLP activation tensors, e.g., $A_{\text{gate}}^{(\ell)}$, $A_{\text{up}}^{(\ell)}$, and $Y_{\text{down}}^{(\ell)}$ for Llama-family models \citep{grattafiori2024llama3herdmodels}.

\begin{figure}[!tb]
  \centering
  \vspace{-4pt}
  \includegraphics[width=\columnwidth]{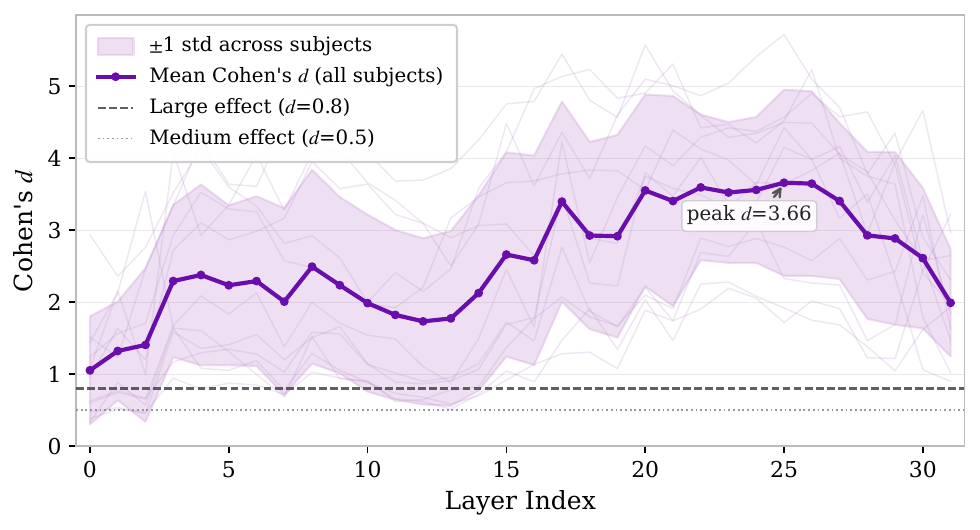}
  \vspace{-6pt}
  \caption{\textbf{Mean Cohen's $d$ across transformer layers on Llama-3.1-8B.} Subject-specific activation separability is strongest in mid-to-late layers, motivating capsule placement in high-salience layers.}
  \label{fig:combined_cohens_d}
\end{figure}

We extract subject-specific signatures through contrastive analysis \citep{rimsky-etal-2024-steering} over token-averaged, standardized activations. Positives come from on-topic subject prompts and negatives from real controls plus prompts from other dataset subjects. These cross-subject negatives are genuine activations for semantically related entity queries, forcing the mined direction to distinguish the target subject from other natural entity representations. 
Since signatures are estimated per target subject, the key coverage criterion is per-subject prompt diversity and contrast quality, not total subject count. If the real negative pool is insufficient, we add a bounded synthetic backfill capped at 10\% of the negative pool, used only for robustness rather than as the primary contrastive distribution.

The primary signature direction is the standardized mean difference:
\textbf{$\mathbf{d}=\mathrm{mean}(S_{\text{pos}}) - \mathrm{mean}(S_{\text{neg}})$},
where $S_{\text{pos}}$ and $S_{\text{neg}}$ denote positive and negative activation sets after standardization. The vector captures the dominant displacement induced by subject-conditioned prompts relative to the matched negative pool. 
We validate localization by measuring layer-wise Cohen's $d$ \citep{Cohen_2013} per subject of the dataset; Figure~\ref{fig:combined_cohens_d} shows strong separability in mid-to-late layers. We use this as a localization diagnostic for finding suitable candidates for capsule placement in Section~\ref{sec:cap_forge}.
\subsection{Capsule-based Suppression}
\label{sec:cap_forge}

\begin{figure*}[t]
  \includegraphics[width=1\linewidth]{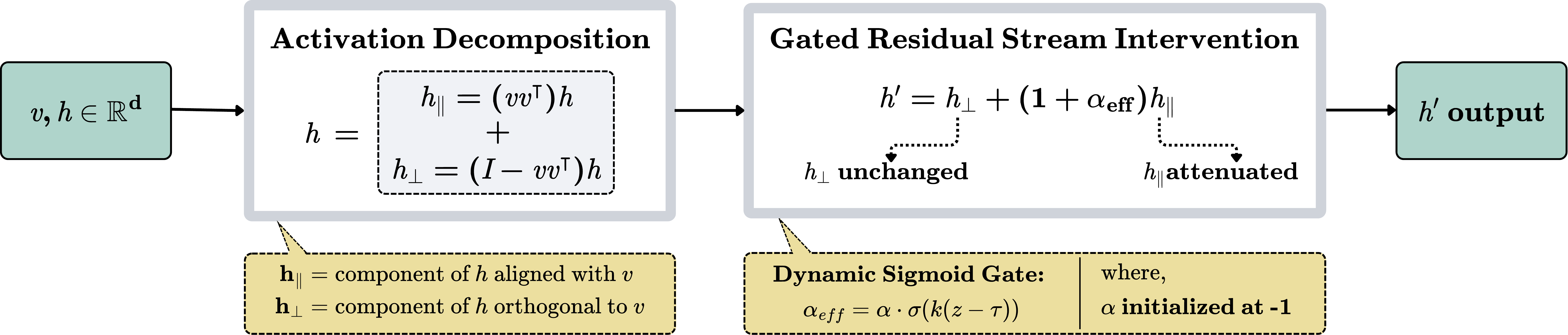}
  \caption{\textbf{Capsule-based suppression.} The capsule decomposes an activation into components parallel and orthogonal to the subject signature. A gated suppression factor is applied only to the parallel component before the modified activation is passed to the subsequent layer.}
  \label{fig:cf}
\end{figure*}

Next, we use the \textbf{Signature Suppression Capsule} (Figure \ref{fig:cf}): a lightweight geometric adapter inspired by activation modification \citep{stoehr-etal-2024-activation, wen-etal-2025-lock} and rank-one editing \citep{meng2023locatingeditingfactualassociations}, to suppress targeted representations at inference while preserving unrelated knowledge and model performance.

For hidden state $h$ and unit signature vector $v$, the capsule decomposes $h$ into $h_{\parallel}$ and $h_{\perp}$. It scales $h_{\parallel}$ through a dynamic attenuation factor $\alpha_{\text{eff}}$ while preserving $h_{\perp}$, thereby targeting the subject-linked direction without altering orthogonal information.

\paragraph{Gated Residual Stream Intervention.}
The attenuation multiplier is $\alpha_{\text{eff}}=\alpha\cdot\sigma(k(z-\tau))$, with $\alpha$ initialized to $-1$. The gate triggers when the state's projection onto the signature, standardized as a Z-score ($z$), exceeds threshold $\tau$ under gain $k$. This mechanism limits intervention to states with statistically elevated alignment to the subject signature. The capsule is implemented through forward hooks with $v$ registered as a fixed buffer. It can be injected dynamically during the UPU Loop (Section \ref{sec:loop}), or removed once unlearning is complete.
\vspace{-1.5pt}

\subsection{Distillation through the Utility Preserving Unlearning Loop}
\label{sec:loop}
Capsules provide immediate suppression, but they are not themselves a permanent parameter update. We therefore distill capsule behavior into a global LoRA adapter through a \textbf{Utility Preserving Unlearning Loop (UPU Loop)}. Whenever the capsule gate triggers at layer $\ell$, we log a tuple $(x,y^+,y^-)$ containing the prompt, the capsule-suppressed response, and the base model's factual output. LoRA parameters $\phi$ are optimized with the following composite objective $\mathcal{L}$ (Eq.~\ref{L}):

\begin{equation}
\small
\begin{split}
\mathcal{L}_{\mathrm{DPO}} = -\mathbb{E}_{\mathcal{D}_{\mathrm{pref}}} \Bigg[ w \cdot \log \sigma \Bigg( & \beta \log \frac{p_\theta(y^+|x)}{p_{\mathrm{ref}}(y^+|x)} \\
& \hspace{-7px} - \beta \log \frac{p_\theta(y^-|x)}{p_{\mathrm{ref}}(y^-|x)} \Bigg) \Bigg]
\end{split}
\label{ldpo}
\end{equation}
\vspace{-5pt}
\begin{equation}
\small
\mathcal{L}_{\mathrm{NTUL}} = -\frac{1}{T} \sum_{t} \log \Big( 1 - \sum_{v \in V_{name}} p_\theta(v | x, y_{<t}^+) \Big)
\label{lntul}
\end{equation}
\begin{equation}
\small
\begin{split}
\mathcal{L} = \mathcal{L}_{\mathrm{DPO}} &+ \lambda_{\mathrm{UL}}\mathcal{L}_{\mathrm{UL}} + \lambda_{\mathrm{NTUL}}\mathcal{L}_{\mathrm{NTUL}} \\
&+ \lambda_{\mathrm{KL}}\mathcal{L}_{\mathrm{KL}} + \lambda_{\mathrm{EWC}}\mathcal{L}_{\mathrm{EWC}} .
\end{split}
\label{L}
\end{equation}

Equation~\ref{ldpo} applies Direct Preference Optimization \citep{rafailov2024directpreferenceoptimizationlanguage}  to the logged tuples, with penalty scaling factor $w$ anchoring updates to the reference model. Factual Unlikelihood $\mathcal{L}_{\mathrm{UL}}$ penalizes the base factual output $y^-$. Name-Token Unlikelihood $\mathcal{L}_{\mathrm{NTUL}}$ penalizes aggregate probability mass over subject-name tokens $V_{name}$ to reduce soft leakage. Stability is encouraged with KL regularization ($\mathcal{L}_{KL}$) \citep{schulman2017proximalpolicyoptimizationalgorithms,rafailov2024directpreferenceoptimizationlanguage} and Elastic Weight Consolidation ($\mathcal{L}_{EWC}$) \citep{Kirkpatrick_2017} on benign anchor prompts, penalizing shifts in parameters identified as critical.

\section{Experiments and Results}
\label{sec:experiments}
We evaluate ERUF along two axes. The experiments on Real-World Entity Dataset are mechanistic analysis of entity-level unlearning through surface leakage, internal target activation, adversarial recovery, hidden-state drift, and sequential stability. TOFU \texttt{forget10} \citep{maini2024tofu} provides standardized comparison against prior unlearning baselines.

\paragraph{Model and baseline selection.}
We evaluate standard foundation models (Llama~\citep{grattafiori2024llama3herdmodels}, Mistral~\citep{jiang2023mistral}) and reasoning-prior models (Qwen~\citep{yang2025qwen3technicalreport}, DeepSeek~\citep{bi2024deepseek}) from 3B to 32B parameters. For TOFU, we compare against Gradient Ascent~\citep{jang-etal-2023-knowledge}, GradDiff~\citep{maini2024tofu}, IDK/refusal~\citep{maini2024tofu}, NPO~\citep{zhang2024negativepreferenceoptimizationcatastrophic}, and SimNPO~\citep{fan2024simplicity}. For entity-level robustness, we also include OPT-OUT where available. Baseline TOFU numbers are from \citet{fan2024simplicity}. Implementation details are provided in Appendix~\ref{sec:tofu_protocol}.

\paragraph{Evaluation metrics.}
For Real-World Entity experiments, we report: \textbf{Utility Drift}, measured as relative benign perplexity change; \textbf{Subject Mention Rate (SMR)}, the percentage of prompts whose generation contains the target name; and \textbf{EL10 ratio}, an early-step subject-associated token probability-mass ratio relative to the base model. EL10 uses up to 10 subject-associated single-token keywords, with canonical subject-name subtokens as backfill, and reports the post/pre probability-mass ratio over the first 10 decoding steps. EL10 is a proxy for latent target activation; it is not a proof of irrecoverability. Following the unlearning-vs.-obfuscation distinction \citep{sun2025unlearningobfuscation}, we use three descriptive mechanism states with tolerance $\epsilon=5\%$:
\begin{itemize}[itemsep=0pt, parsep=1.5pt, topsep=0pt]
    \item \textbf{Type I: representation-level attenuation.} SMR $\le \epsilon$ and EL10 $<1$.
    \item \textbf{Type II: obfuscation.} SMR $\le \epsilon$ and EL10 $>1$.
    \item \textbf{Type III: instability.} SMR $>\epsilon$.
\end{itemize}
For TOFU, we report Forget Quality (FQ) and Model Utility (MU) \citep{maini2024tofu}.

\begin{table*}[htpb]
\centering
\small
\begin{tabular}{llcccl}
\toprule
\textbf{Family} & \textbf{Model Size} & \textbf{Util. Drift} & \textbf{SMR} ($\downarrow$) & \textbf{EL10} ($\downarrow$) & \textbf{Mechanism State} \\
\midrule
Standard & Mistral 7B & $+0.50\% \pm 0.45$ & $0.00\% \pm 0.00$ & $0.020 \pm 0.015$ & Type I \\
Standard & Llama 8B & $-0.45\% \pm 1.59$ & $1.10\% \pm 1.90$ & $0.052 \pm 0.047$ & Type I \\
Standard & Mistral 3B & $+0.36\% \pm 2.27$ & $0.52\% \pm 0.35$ & $0.054 \pm 0.047$ & Type I \\
Reasoning & Qwen 14B & $+1.99\% \pm 0.51$ & $0.52\% \pm 0.45$ & $0.035 \pm 0.013$ & Type I \\
Reasoning & Qwen 32B & $+1.31\% \pm 0.38$ & $0.08\% \pm 0.06$ & $0.024 \pm 0.018$ & Type I \\
\midrule
Reasoning & Qwen 8B & $-0.50\% \pm 0.60$ & $3.33\% \pm 1.25$ & $11.03 \pm 2.45$ & Type II \\
Reasoning & DeepSeek 8B & $+1.03\% \pm 0.85$ & $0.00\% \pm 0.00$ & $6.19 \pm 1.80$ & Type II \\
Standard & Llama 3B & $-0.44\% \pm 0.40$ & $0.00\% \pm 0.00$ & $3.06 \pm 0.55$ & Type II \\
\midrule
Reasoning & Qwen 3B & $+2.22\% \pm 2.42$ & $0.43\% \pm 0.37$ & $1.460 \pm 1.480$ & Type II \\
Reasoning & DeepSeek 3B & $+2.15\% \pm 1.74$ & $45.6\% \pm 33.6$ & $15.39 \pm 15.54$ & Type III \\
\midrule
Baseline & LUNAR-Llama 8B & $+32.80\%$ & $56.06\%$ & $7.57$ & Type III \\
Baseline & ReGLU-Llama 8B & $+51.34\%$ & $1.52\%$ & $13.13$ & Type II \\
Baseline & OPT-OUT-Llama 8B & $+1.31\%$ & $50.00\%$ & $1.62$ & Type III \\
Baseline & SimNPO-Llama 8B & $+1.23\%$ & $72.73\%$ & $0.82$ & Type III \\
\bottomrule
\end{tabular}
\caption{\textbf{Cross-model mechanistic validation.} Results are mean$\pm$SD across three seeds where available; baseline rows are reported as single evaluated conditions. ERUF achieves Type-I behavior on most standard models and on Qwen 14B/32B, while exposing Type-II obfuscation in Llama 3B and 8B reasoning-prior models and Type-III instability in small reasoning-prior models. Evaluated baselines show either high leakage (Type III) or elevated EL10 (Type II) for ReGLU, indicating weaker mechanistic removal.}

\label{tab:cross_model_mechanistic_validation}
\end{table*}

\begin{figure}[!tb]
  \centering
  \includegraphics[width=\columnwidth]{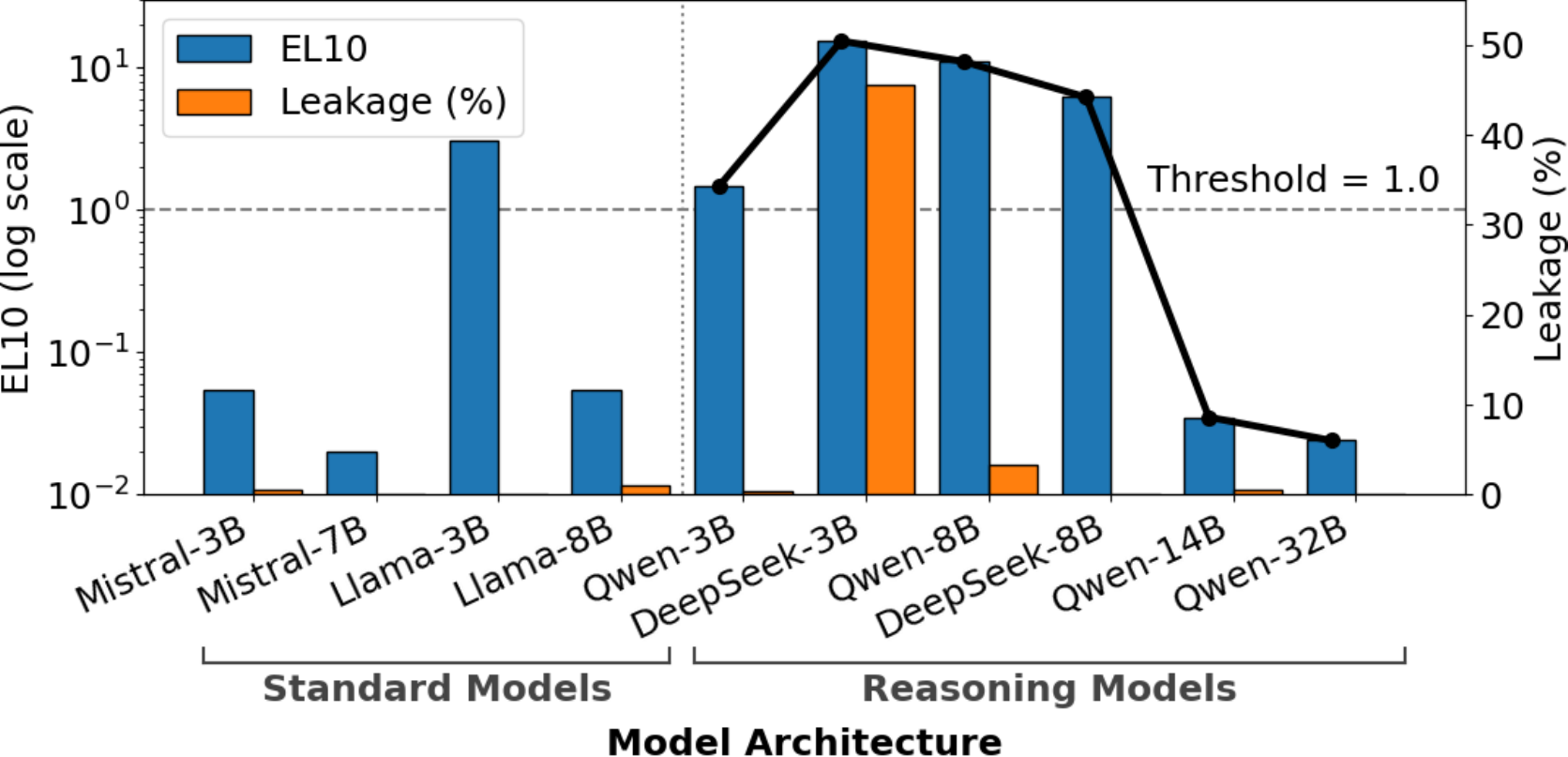}
  \caption{\textbf{EL10 and surface leakage across model families and scales.} Standard models mostly maintain low internal target activation, with Llama 3B as an exception. Reasoning-prior models show a capacity-dependent pattern.}
  \label{fig:el10_standard}
\end{figure}

\subsection{Mechanism States on the Real-World Entity Dataset}
\label{realworld_results}


Table~\ref{tab:cross_model_mechanistic_validation} summarizes the main behavioral comparison. ERUF is the only evaluated method that jointly achieves low surface leakage, internal attenuation, and utility preservation on the primary standard-model setting, while the same diagnostic exposes exceptions such as Llama 3B. Baselines expose complementary failure modes: LUNAR~\citep{shen2026llm}, OPT-OUT~\citep{choi-etal-2025-opt}, and SimNPO~\citep{fan2024simplicity} retain high surface leakage, while ReGLU suppresses SMR but, like LUNAR, amplifies EL10 and degrades utility. ERUF's claim, however, is not architecture-uniform: Llama 3B remains Type II, with zero SMR but elevated EL10.

Reasoning-prior models exhibit a capacity-dependent pattern. DeepSeek 3B is unstable (Type-III), Qwen/DeepSeek 8B suppress surface leakage but retain high EL10 (Type-II), and only at 14B parameters does Qwen return to Type-I behavior, which remains consistent at 32B, demonstrating that as parameter count increase, unlearning stabilizes. Thus, SMR alone misses a key transition. Figure~\ref{fig:el10_standard} visualizes why: Qwen 8B and Qwen 14B both have low SMR, but differ by over two orders of magnitude in EL10.


\vspace{-3pt}

\subsection{Adversarial and Name-Agnostic Recovery Diagnostics}\label{sec:adv_recovery}

Output-level unlearning can fail when the canonical name is removed from the prompt. We therefore evaluate ERUF, the pre-unlearning base model, OPT-OUT, SimNPO, ReGLU, and LUNAR on adversarial and name-agnostic recovery probes using Llama-3.1-8B. Since the comparison systems are not specifically adversarially tuned, we treat them as entity-level baselines rather than state-of-the-art adversarial defenses.


We use adversarial recovery diagnostics as benign entity-unlearning stress tests to assess whether forgotten-subject information remains recoverable through indirect or rephrased queries. Following RWKU’s adversarial probes and locality tests \citep{jin2024rwku}, and motivated by evidence that edited or deleted subject information can persist in hidden-state traces or be recovered via rephrased questions \citep{patil2023sensitiveinformationdeletedllms}, we evaluate direct-profile, alias-only, descriptor-only, relation-clue, masked-name, instruction-pressure, multi-turn, multilingual, and mixed-query probes. These test whether entity knowledge survives beyond canonical-name suppression, consistent with findings that forget-set coverage strongly influences entity-unlearning success \citep{ma2025unveilingentitylevelunlearninglarge}. Full attack probe definitions are provided in Appendix~\ref{app:adversarial_recovery}.

\begin{table}[!t]\centering\footnotesize\setlength{\tabcolsep}{3pt}
\renewcommand{\arraystretch}{1.2}
\begin{adjustbox}{max width=\columnwidth}
\begin{tabular}{@{}lccccc@{}}
\toprule
\textbf{Model}
  & \shortstack{\textbf{Alias}\\\textbf{Hit\%}$\downarrow$}
  & \shortstack{\textbf{Keyword}\\\textbf{Hit\%}$\downarrow$}
  & \shortstack{\textbf{Target}\\\textbf{Mass}$\downarrow$}
  & \shortstack{\textbf{Recovery\%}\\$\downarrow$} \\
\midrule
\textsc{Pre}     & 31.95 & 16.53 & $4.398{\times}10^{-2}$ & 63.89 \\
\textsc{Opt-Out} & 21.78 & 13.57 & $2.874{\times}10^{-2}$ & 54.41 \\
\textsc{SimNPO}  & 36.23 & 21.46 & $3.999{\times}10^{-2}$ & 74.63 \\
\textsc{ReGLU}   &  9.57 & 18.28 & $3.596{\times}10^{-2}$ & 72.12 \\
\textsc{LUNAR}   & 33.99 & 16.08 & $4.096{\times}10^{-2}$ & 70.22 \\
\textsc{ERUF}     & \textbf{9.54} & \textbf{4.63} & \textbf{$1.357{\times}10^{-2}$} & \textbf{20.15} \\
\bottomrule
\end{tabular}
\end{adjustbox}
\vspace{4pt}
\caption{\textbf{Overall adversarial forget-recovery results.} Target mass is reported as mean probability mass. ERUF has the lowest recovery rate and target-token mass among the evaluated conditions. Comparison systems are used as entity-level baselines rather than adversarially tuned state-of-the-art methods.}

\label{tab:overall_adv}
\end{table}

\begin{table*}[!t]\centering\scriptsize\setlength{\tabcolsep}{2.5pt}
\renewcommand{\arraystretch}{1.05}
\begin{adjustbox}{max width=\textwidth}
\begin{tabular}{@{}lcccccc cccccc@{}}
\toprule
 & \multicolumn{6}{c}{\textbf{Recovery Success Percentage}$\downarrow$}
 & \multicolumn{6}{c}{\textbf{Alias Hit Percentage}$\downarrow$} \\
\cmidrule(lr){2-7}\cmidrule(lr){8-13}
\textbf{Attack Family}
  & \textsc{Pre} & \textsc{Opt-Out} & \textsc{SimNPO} & \textsc{ReGLU} & \textsc{LUNAR} & \textsc{ERUF}
  & \textsc{Pre} & \textsc{Opt-Out} & \textsc{SimNPO} & \textsc{ReGLU} & \textsc{LUNAR} & \textsc{ERUF} \\
\midrule
Alias-only           & 88.89 & 73.94 & 87.88 & 93.94 & 69.70 & \textbf{22.47} & 57.78 & 37.58 & 55.76 & 36.97 & 38.18 & \textbf{7.87} \\
Descriptor-only      & 69.71 & 52.53 & 85.86 & 81.82 & 71.21 & \textbf{21.26} & 18.75 & 10.10 & 25.25 &  4.04 & 19.19 & \textbf{9.66} \\
Direct profile       & 77.14 & 72.73 & 84.85 & 54.55 & 78.79 & \textbf{12.86} & 61.43 & 54.55 & 65.15 & 16.67 & 56.06 & \textbf{1.43} \\
Instruction pressure & 69.73 & 59.39 & 70.91 & 59.39 & 72.73 & \textbf{34.05} & 42.16 & 35.15 & 40.61 &  4.85 & 49.70 & \textbf{24.86} \\
Masked name          & 40.00 & 45.45 & 38.18 & 63.64 & 49.09 & \textbf{21.67} & 16.67 &  7.27 & 10.91 &  3.64 & 21.82 & \textbf{5.00} \\
Mixed query          & 39.75 & 25.00 & 65.00 & 68.18 & 65.45 &  \textbf{5.04} & 15.48 &  3.18 & 16.82 &  6.82 & 23.18 & \textbf{0.00} \\
Multi-turn style     & 41.85 & 63.03 & 51.52 & 55.76 & 81.82 & \textbf{23.50} & 23.37 & 18.79 & 26.67 &  4.85 & 39.39 & \textbf{14.21} \\
Multilingual simple  & 84.46 & 71.02 & 93.75 & 85.23 & 80.11 & \textbf{32.81} & 38.34 & 35.23 & 52.84 &  4.55 & 44.89 & \textbf{11.98} \\
Relation clue        & 79.00 & 55.05 & 90.91 & 72.73 & 68.69 & \textbf{15.53} & 37.90 & 18.69 & 45.96 &  4.55 & 28.79 & \textbf{9.13} \\
\bottomrule
\end{tabular}
\end{adjustbox}

\caption{\textbf{Adversarial forget-recovery by attack family.}
ERUF has the lowest recovery success in every reported attack family. Alias-hit results show that reduced recovery is not limited to exact-name suppression.}
\label{tab:adv_family}
\end{table*}

Tables~\ref{tab:overall_adv}--\ref{tab:adv_family} show that ERUF is consistently more robust under adversarial and name-agnostic probes than the evaluated baselines. The strongest evidence comes from probes where the canonical name is absent or indirect, which tests whether forgetting extends beyond exact-name refusal. These results are consistent with attenuation of non-canonical access routes rather than simple surface suppression. However, residual recovery under harder prompt families like multilingual, multi-turn, and instruction-pressure prompts, shows that recover-ability is reduced rather than eliminated.


\begin{table}[!t]\centering\footnotesize\setlength{\tabcolsep}{2pt}
\renewcommand{\arraystretch}{1.1}
\begin{tabular}{@{}lccccc@{}}
\toprule
\textbf{Model}
  & \shortstack{\textbf{N-agn.}\\\textbf{alias}$\downarrow$}
  & \shortstack{\textbf{N-agn.}\\\textbf{kwd.}$\downarrow$}
  & \shortstack{\textbf{N-agn.}\\\textbf{mass}$\downarrow$}
  & \shortstack{\textbf{Mxd.}\\\textbf{alias}$\downarrow$}
  & \shortstack{\textbf{Mxd.}\\\textbf{mass}$\downarrow$} \\
\midrule
\textsc{Pre}     & 0.311 & 0.179 & 0.0423 & 0.0180 & 0.0142 \\
\textsc{Opt-Out} & 0.318 & 0.179 & 0.0418 & 0.0180 & 0.0141 \\
\textsc{SimNPO}  & 0.353 & 0.213 & 0.0374 & 0.0293 & 0.0175 \\
\textsc{ReGLU}   & 0.152 & 0.193 & 0.0369 & 0.0313 & 0.0138 \\
\textsc{LUNAR}   & 0.267 & 0.141 & 0.0332 & 0.0703 & 0.0218 \\
\textsc{ERUF}     & $\mathbf{0.070}$ & $\mathbf{0.0469}$ & $\mathbf{0.0115}$ & $\mathbf{0.00719}$ & $\mathbf{0.00469}$ \\
\midrule
\textit{ERUF red.} & \textit{77.4\%} & \textit{73.8\%} & \textit{72.7\%} & \textit{60.0\%} & \textit{66.9\%} \\
\bottomrule
\end{tabular}
\vspace{1.5pt}
\caption{\textbf{Name-agnostic and mixed-query forget robustness.} N-agn.\ = name-agnostic; Mxd.\ = mixed-query; kwd.\ = keyword hit; mass = target mass. All results use the same 750 evaluated row identifiers. ERUF reduction (last row) is relative to Pre.}
\label{tab:name_agnostic}
\end{table}

Table~\ref{tab:name_agnostic} shows that ERUF strongly reduces alias, keyword, and target-mass diagnostics when the canonical subject name is removed, while several baselines remain close to the pre-unlearning model. This supports the interpretation that ERUF weakens non-canonical access routes.

\vspace{-3pt}
\subsection{Hidden-State Selectivity}
\label{sec:hidden_main}
We examine whether ERUF shifts forgotten-subject representations more than unrelated prompts. Table~\ref{tab:hidden_space_diag_main} summarizes hidden-space diagnostics on Llama-3.1-8B. Relative to the pre-unlearning base model, ERUF achieves zero surface leakage, a 95.6\% reduction in EL10 mass, and a 30.8\% reduction in $E_{30}$ extraction mass. It also produces the largest forget-prompt drift and the highest selective representation shift ratio (SRS), indicating that forget-subject representations move substantially more than benign representations. Baselines either retain target exposure or show weaker forget-vs-benign selectivity. This suggests that ERUF changes forgotten-subject representations more selectively than output-only suppression would imply.

\begin{table}[t]
\centering
\small
\setlength{\tabcolsep}{2.6pt}
\renewcommand{\arraystretch}{1.12}
\begin{adjustbox}{max width=\columnwidth}
\begin{tabular}{@{}lcccccc@{}}
\toprule
\textbf{Model}
& \textbf{SMR} $\downarrow$
& \textbf{EL10 mass} $\downarrow$
& $\boldsymbol{D_{\ell}^{F}}$ $\uparrow$
& $\boldsymbol{D_{\ell}^{B}}$ $\downarrow$
& \textbf{SRS} $\uparrow$
& $\boldsymbol{E_{30}}$ $\downarrow$ \\
\midrule
Pre base
& $6.67{\times}10^{-1}$
& $1.92{\times}10^{-3}$
& --
& --
& --
& $1.26{\times}10^{-4}$ \\
\textsc{Opt-Out}
& $4.85{\times}10^{-1}$
& $4.59{\times}10^{-2}$
& $7.00{\times}10^{-4}$
& $\mathbf{5.00{\times}10^{-4}}$
& $1.30{\times}10^{0}$
& $3.10{\times}10^{-2}$ \\
\textsc{LUNAR}
& $7.27{\times}10^{-1}$
& $6.44{\times}10^{-2}$
& $0.00{\times}10^{0}$
& $0.00{\times}10^{0}$
& $0.00{\times}10^{0}$
& $6.36{\times}10^{-2}$ \\
\textsc{ReGLU}
& $1.06{\times}10^{-1}$
& $1.01{\times}10^{-2}$
& $3.40{\times}10^{-3}$
& $3.60{\times}10^{-3}$
& $9.00{\times}10^{-1}$
& $4.73{\times}10^{-3}$ \\
\textsc{SimNPO}
& $7.73{\times}10^{-1}$
& $4.07{\times}10^{-2}$
& $0.00{\times}10^{0}$
& $0.00{\times}10^{0}$
& $1.30{\times}10^{0}$
& $4.00{\times}10^{-2}$ \\
ERUF
& $\mathbf{0.00{\times}10^{0}}$
& $\mathbf{8.38{\times}10^{-5}}$
& $\mathbf{4.25{\times}10^{-2}}$
& $7.17{\times}10^{-3}$
& $\mathbf{5.90{\times}10^{0}}$
& $\mathbf{8.72{\times}10^{-5}}$ \\
\bottomrule
\end{tabular}%
\end{adjustbox}

\caption{\textbf{Hidden-space diagnostics on Llama-3.1-8B.}
$D_{\ell}^{F}$ and $D_{\ell}^{B}$ denote hidden-state drift at layer $\ell$ on forget prompts and generic benign prompts, respectively, measured against the pre-unlearning base model. SRS is the selective representation shift ratio $D_{\ell}^{F}/(D_{\ell}^{B}+\epsilon)$, computed from unrounded drift values; printed zero drift values indicate values below display precision. $E_{30}$ denotes subject-associated token extraction mass over a 30-step autoregressive rollout.}

\label{tab:hidden_space_diag_main}
\end{table}

This hidden-state evidence addresses a limitation of output-only evaluation. A refusal-only mechanism could alter generated text without selectively shifting hidden states for forgotten-subject prompts. ERUF does shift benign representations, but its forget-prompt drift is 5.9$\times$ larger than its benign drift, suggesting target-directed rather than global representation change.

\vspace{-3pt}
\subsection{Standardized and Auxiliary Evaluations}
\paragraph{TOFU Benchmark Results}
\label{tofu_results}

\begin{table}[!tb]
  \centering
  \footnotesize
  \setlength{\tabcolsep}{2.8pt}
  \renewcommand{\arraystretch}{1.02}
  \resizebox{\columnwidth}{!}{%
  \begin{tabular}{@{}lcccccccc@{}}
    \toprule
    \textbf{Metric}
      & \textbf{Oracle}
      & \textbf{Orig.}
      & \textbf{GA}
      & \textbf{GD}
      & \textbf{IDK}
      & \textbf{NPO}
      & \textbf{Sim}
      & \textbf{ERUF} \\
    \midrule
    FQ $\uparrow$
      & $1.00$
      & $0.00$
      & $2.2{\times}10^{-16}$
      & $3.7{\times}10^{-15}$
      & $2.9{\times}10^{-14}$
      & $0.29$
      & $0.45$
      & $\mathbf{0.99}$ \\
    MU $\uparrow$
      & $0.62$
      & $0.62$
      & $0.00$
      & $0.54$
      & $0.54$
      & $0.55$
      & $0.62$
      & $\mathbf{0.62}$ \\
    \bottomrule
  \end{tabular}%
  }
  \vspace{2pt}
  \caption{\textbf{TOFU-\texttt{forget10} on LLaMA-2-7B-Chat.}
  We report Forget Quality (FQ) and Model Utility (MU). Baselines are from \citet{fan2024simplicity} (Table A3). ERUF achieves near-oracle FQ while matching original/oracle MU. Abbreviations: GA = Gradient Ascent, GD = GradDiff, Sim = SimNPO.}
  
  \label{tab:tofu_comparison}
\end{table}

Table~\ref{tab:tofu_comparison} reports the standardized TOFU \texttt{forget10} result. ERUF approaches oracle-level forgetting while preserving aggregate utility at the original-model level. Compared with prior baselines, this places ERUF on a more favorable forgetting--utility trade-off. This result is computed on the TOFU-origin LLaMA-2-7B-Chat checkpoint for comparability with prior work.

\vspace{-3pt}
\paragraph{RWKU Evaluation Results}
\label{sec:rwku}
To test whether mined signatures generalize beyond our custom prompt templates,
we conduct an evaluation-only probe on the four subjects shared between our
dataset and RWKU~\citep{jin2024rwku} on Llama-3.1-8B. Table~\ref{tab:rwku_combined}
shows consistent post-unlearning reductions across FB, QA, and adversarial-attack
recall, along with an increase in FM loss, confirming that the mined signature
is a faithful subject-level representation whose suppression generalizes across
independently-constructed probe formats. Because the evaluation uses only the
four subjects shared between our dataset and RWKU, it should be interpreted as
a restricted generalization probe rather than a full RWKU leaderboard comparison. 



\begin{table}[t]
\centering
\small
\setlength{\tabcolsep}{3.5pt}
\renewcommand{\arraystretch}{1.05}
\begin{adjustbox}{max width=\columnwidth}
\begin{tabular}{lcccccccc}
\toprule
& \multicolumn{2}{c}{\textbf{FB} ($\downarrow$)}
& \multicolumn{2}{c}{\textbf{QA} ($\downarrow$)}
& \multicolumn{2}{c}{\textbf{AA} ($\downarrow$)}
& \multicolumn{2}{c}{\textbf{FM Loss} ($\uparrow$)} \\
\cmidrule(lr){2-3}\cmidrule(lr){4-5}
\cmidrule(lr){6-7}\cmidrule(lr){8-9}
\textbf{Subject}
  & Pre & Post & Pre & Post & Pre & Post & Pre & Post \\
\midrule
Ariana Grande
  & 0.778 & 0.444 & 0.729 & 0.246
  & 0.500 & 0.316 & 1.857 & 2.059 \\
Beyonc\'{e}
  & 0.725 & 0.525 & 0.796 & 0.389
  & 0.567 & 0.183 & 1.838 & 2.006 \\
Kanye West
  & 0.750 & 0.633 & 0.708 & 0.354
  & 0.563 & 0.279 & 2.127 & 2.242 \\
Taylor Swift
  & 0.900 & 0.800 & 0.733 & 0.457
  & 0.526 & 0.346 & 2.042 & 2.224 \\
\midrule
\textbf{Mean}
  & \textbf{0.788} & \textbf{0.601}
  & \textbf{0.742} & \textbf{0.362}
  & \textbf{0.539} & \textbf{0.281}
  & \textbf{1.966} & \textbf{2.133} \\
\bottomrule
\end{tabular}%
\end{adjustbox}
\caption{\textbf{RWKU generalization probe} (4-subject subset; evaluation only). ROUGE-L recall on FB, QA, and AA probes, and FM loss from the MIA set, before and after ERUF. Lower FB/QA/AA and higher FM loss indicate reduced recover-ability.}
\label{tab:rwku_combined}
\end{table}

\begin{table*}[!t]
\centering
\setlength{\tabcolsep}{5pt}
\renewcommand{\arraystretch}{1.08}
\resizebox{\textwidth}{!}{%
\begin{tabular}{lccccccccc}
\toprule
\textbf{Group} & \textbf{n}
  & \textbf{SMR$_\text{Pre}$} & \textbf{SMR$_\text{S1}$} & \textbf{SMR$_\text{S2}$} & $\Delta$
  & \textbf{EL10$_\text{Pre}$} & \textbf{EL10$_\text{S1}$} & \textbf{EL10$_\text{S2}$} & $\Delta$ \\
\midrule
Previous forget targets (Batch~1)
  & 5  & 0.8000 & 0.0000 & 0.0000 & $-$0.8000
  & $3.04{\times}10^{-5}$ & $5.07{\times}10^{-5}$ & $1.55{\times}10^{-6}$ & $-2.88{\times}10^{-5}$ \\
New forget targets (Batch~2)
  & 5  & 0.8000 & 0.5667 & 0.0000 & $-$0.8000
  & $9.93{\times}10^{-3}$ & $9.79{\times}10^{-3}$ & $1.36{\times}10^{-7}$ & $-9.93{\times}10^{-3}$ \\
All subjects
  & 10 & 0.8000 & 0.2833 & 0.0000 & $-$0.8000
  & $4.98{\times}10^{-3}$ & $4.92{\times}10^{-3}$ & $8.44{\times}10^{-7}$ & $-4.98{\times}10^{-3}$ \\
\bottomrule
\end{tabular}%
}

\caption{\textbf{Sequential unlearning summary.} Stage~1 (S1) forgets Batch~1; Stage~2 (S2) adds a disjoint Batch~2 and unlearns it. Previously forgotten subjects do not re-emerge, and new targets are suppressed in the final model. The joint drop in SMR and EL10 indicates sequential stability at both surface and latent-activation levels.}

\label{tab:seq_summary}
\end{table*}

\vspace{-3pt}
\paragraph{General Capability Retention}
\label{sec:capability_retention}
\begin{table}[!tb]
    \centering
    \begin{adjustbox}{max width=\columnwidth}
\begin{tabular}{lcccc}
        \toprule
        \textbf{Benchmark} & \textbf{Metric} & \textbf{Pre (Base)} & \textbf{Post (Adapter)} & \textbf{Delta} \\
        \midrule
        ARC-Challenge      & acc\_char & 76.88 & 76.62 & \textcolor{red}{-0.26} \\
        OpenBookQA         & acc\_char & 74.80 & 74.40 & \textcolor{red}{-0.40} \\
        HellaSwag          & acc       & 76.68 & 76.84 & \textcolor{green}{+0.16} \\
        BoolQ              & acc       & 78.93 & 74.46 & \textcolor{red}{-4.47} \\
        PIQA               & acc\_char & 72.09 & 71.65 & \textcolor{red}{-0.44} \\
        SocialIQA          & acc\_char & 59.88 & 55.99 & \textcolor{red}{-3.89} \\
        WinoGrande         & acc       & 68.98 & 69.46 & \textcolor{green}{+0.48} \\
        TruthfulQA (MC2)   & mc2       & 40.36 & 40.87 & \textcolor{green}{+0.51} \\
        \bottomrule
        \end{tabular}%
\end{adjustbox}
    \caption{\textbf{Zero-shot capability retention.}
Post-unlearning performance remains close to the base model on most benchmarks, with localized degradation on BoolQ and SocialIQA. Positive and negative deltas indicate changes relative to the base model.}
    \label{tab:cap_ret_zero}
\end{table}

We evaluate zero-shot performance on ARC-Challenge~\citep{clark2018think}, OpenBookQA~\citep{mihaylov2018can}, BoolQ~\citep{clark2019boolq}, HellaSwag~\citep{zellers2019hellaswag}, PIQA~\citep{bisk2020piqa}, SocialIQA~\citep{sap-etal-2019-social}, WinoGrande~\citep{10.1145/3474381}, and TruthfulQA~\citep{lin-etal-2022-truthfulqa}. Table~\ref{tab:cap_ret_zero} shows that post-unlearning performance is close to the base model on most benchmarks, with localized degradation on some question-answering and social-reasoning tasks. This supports broad, but not uniform, capability retention.

\vspace{-3pt}
\subsection{Ablation}
\label{sec:ablation}

\begin{table}[!t]
\centering
\small
\begin{adjustbox}{max width=\columnwidth}
\begin{tabular}{lcccl}
\toprule
\textbf{Exp} & \textbf{SMR} ($\downarrow$) & \textbf{EL10} ($\downarrow$) & \textbf{Util. Drift} & \textbf{State} \\
\midrule
A: Full       & 0.00\% & 0.066 & \textcolor{green}{+0.45\%} & Type I \\
B: No NT-UL   & 3.33\% & 1.275 & \textcolor{red}{-1.35\%} & Type II \\
C: No Factual UL  & 0.00\% & 1.098 & \textcolor{red}{-1.29\%} & Type II \\
\bottomrule
\end{tabular}%
\end{adjustbox}
\caption{\textbf{Ablation results on Llama-3.1-8B.}
Removing Name-Token Unlikelihood increases leakage and raises EL10 above 1 (Type II); removing Factual Unlikelihood keeps leakage controlled but also raises EL10 above 1 (Type II).}

\label{tab:ablation}
\end{table}


Table~\ref{tab:ablation} links each objective component to the resulting mechanism state. Removing Name-Token Unlikelihood increases residual surface leakage and raises EL10 above 1, yielding Type-II obfuscation. Removing Factual Unlikelihood also produces Type-II obfuscation, where surface control remains but internal activation persists. The full objective is the only ablation in this table that maintains Type-I behavior by combining surface suppression with internal attenuation.

\vspace{-3pt}

\subsection{Sequential Stability}
\label{sec:seq_main}

Sequential unlearning is evaluated with a two-stage update over disjoint target batches spanning entertainment, politics, science, sports, and technology. Table~\ref{tab:seq_summary} shows that previously forgotten subjects do not re-emerge after the second update, while newly targeted subjects are suppressed later on. Benign utility remains stable after adding the second forget batch: benign loss changes from $4.7342$ to $4.7308$, and benign perplexity changes from $11.377$ to $11.339$. Because Table~\ref{tab:seq_summary} focuses on forget-target metrics, benign loss/perplexity are reported in text rather than as table columns.

\section{Conclusion}
We presented ERUF, a representation-aware entity unlearning framework that mines subject-specific activation signatures, suppresses them through capsules, and distills the suppressed behavior into LoRA parameters. The main empirical contribution is a behavioral profile (not a claim of novel primitives): among the evaluated baselines, only ERUF simultaneously achieves low surface leakage, reduced internal target activation, and low utility drift on the primary standard-model setting.

The dual metric of SMR and EL10 exposes failure modes that output-only evaluation would miss. ReGLU suppresses the surface while increasing internal activation; SimNPO lowers EL10 but leaks; ERUF itself shows a Llama-3B Type-II exception and reasoning-model Type-II behavior at 8B. Adversarial, name-agnostic, hidden-state, RWKU-subset, and sequential evaluations provide convergent evidence that ERUF attenuates target representations beyond exact-name refusal, while residual recovery and locality caveats prevent claims of complete or mathematically irreversible deletion.

Overall, the results support representation-level diagnostics as a necessary complement to output-level unlearning metrics, especially for entity-level deletion where aliases, clues, and latent reasoning paths can preserve recoverability.

\section*{Limitations}

ERUF provides operational evidence of representation-level attenuation, not a formal guarantee of irrecoverability. EL10, hidden-state drift, target-token mass, and adversarial recovery are proxies; extreme adversarial fine-tuning, future jailbreak methods, or broader prompt searches may recover additional information. We therefore avoid claiming complete deletion in the database sense.

The activation-signature pipeline uses a simple standardized mean-difference direction and relies on a matched negative pool rather than a learned localization model. Because same-subject controls are sparse, most negatives are real cross-subject entity activations, with bounded synthetic backfill used only when the real negative pool is insufficient. This makes mining robust across subjects, but Cohen's $d$ should still be interpreted as a relative signature-separability diagnostic for candidate layer selection, not as absolute discriminability against all possible natural negative distributions. The Real-World Entity Dataset is imbalanced across subjects, so per-subject variation remains possible.

Our resource constraints limited experiments to 4-bit quantized models up to 32B parameters on a single RTX A6000. The TOFU result is a single run for direct comparability with published baselines; seed variance is evaluated primarily on the Real-World Entity Dataset and capsule gate sweep. The RWKU evaluation uses only four shared subjects and is not a full leaderboard comparison. Finally, while standard utility benchmarks are mostly stable, same-domain locality stress tests show that ERUF can suppress related non-forgotten music entities, so the method should not be interpreted as perfect entity-local editing.

\section*{Ethical Consideration}
\label{sec:ethics}
ERUF has dual-use implications. It can support privacy compliance, including GDPR-related erasure requests \citep{gdpr2016, googlespain2014, zhang2024rightforgotten}, and the reduction of sensitive or copyrighted memorization. However, targeted erasure mechanisms could also be misused for censorship, partisan manipulation, or guardrail removal \citep{Weidinger2021EthicalAS, Zou2023UniversalAT, Zhao2024WeaktoStrongJO}. Because neural unlearning does not provide the same certainty as deleting a database record, users should not treat reduced leakage as a guarantee that information is unrecoverable \citep{9519428,sun2025unlearningobfuscation,Zou2023UniversalAT,Zhao2024WeaktoStrongJO}.

Targeted erasure may also create semantic holes or uneven effects on related concepts. Our locality caveats show that same-domain related entities can be affected under some robustness probes. Deployments should therefore include auditing for demographic, topical, and domain-specific collateral damage. From a sustainability perspective, ERUF uses PEFT and avoids full retraining, reducing compute and associated carbon cost relative to retraining-based deletion \citep{strubell-etal-2019-energy,Patterson2021CarbonEA}. We acknowledge the use of generative AI models for creating the conceptual visualization in Figure~\ref{fig:motivation} and for improving clarity and grammar in the manuscript. 

All external artifacts used in this paper, including benchmarks, codebases, and databases, are collected from resources publicly made available by their respective authors and are used only for their intended purposes and within the terms of the provided licenses; baseline codebases follow the authors’ publicly available GitHub repositories, with hyperparameters set according to the corresponding papers.

\bibliography{custom}

\clearpage
\appendix

\section{Dataset Construction}
\label{sec:dataset}
We construct a dataset of controlled prompts grounded in verifiable facts about real-world entities. We select 11 public subjects, and for each subject we extract facts from Wikipedia and Wikidata. From Wikipedia, we store a short description from the article summary, selected infobox key--value pairs, and limited section snippets. From Wikidata, we query high-salience properties using SPARQL and store the property label as the predicate and the resolved label or value as the object. Each factual triple record includes a deterministic ID and source URL for traceability.

\begin{table}[H]
\centering
\small
\setlength{\tabcolsep}{6pt}
\renewcommand{\arraystretch}{1.15}
\begin{tabularx}{\columnwidth}{@{}l l X@{}}
\toprule
\textbf{Subject} & \textbf{Predicate} & \textbf{Example object} \\
\midrule
Beyonc\'{e} & award received & Grammy Award for Song of the Year \\
Beyonc\'{e} & occupation & singer \\
Beyonc\'{e} & nominated for & Academy Award for Best Original Song \\
\midrule
Kanye West & occupation & rapper \\
Kanye West & nominated for & Grammy Award for Record of the Year \\
Kanye West & award received & Grammy Award for Best Rap Song \\
\midrule
Taylor Swift & award received & Grammy Award for Album of the Year \\
Taylor Swift & occupation & singer \\
Taylor Swift & notable work & Shake It Off \\
\bottomrule
\end{tabularx}
\caption{Example subjects with three predicate--object pairs per subject.}
\label{tab:top-subjects-top-predicates}
\end{table}
\FloatBarrier

\subsection{Dataset Statistics}
The dataset contains $5{,}824$ total prompts instantiated from $604$ triples mined for $11$ subjects, with five templates per category. Table~\ref{tab:prompt_categories} shows prompt counts by category, and Table~\ref{tab:triples_per_subject} shows triples per subject.

\begin{table}[h]
\centering
\begin{tabular}{|l|r|}
\hline
\textbf{Prompt Category} & \textbf{Count} \\ \hline
Implicit & 1360 \\
Direct & 2416 \\
Contextual & 1184 \\
Reasoning & 132 \\
Misleading & 732 \\ \hline
\end{tabular}
\caption{Distribution of prompt categories.}
\label{tab:prompt_categories}
\end{table}

\begin{table}[h]
\centering
\begin{tabular}{|l|r|}
\hline
\textbf{Subject} & \textbf{Count} \\ \hline
Ariana Grande & 57 \\
Arijit Singh & 37 \\
Beyonc\'{e} & 109 \\
Drake (musician) & 14 \\
Ed Sheeran & 61 \\
Eminem & 14 \\
Kanye West & 95 \\
Katy Perry & 75 \\
Michael Jackson & 35 \\
Queen (band) & 12 \\
Taylor Swift & 95 \\ \hline
\end{tabular}
\caption{Knowledge triples per subject.}
\label{tab:triples_per_subject}
\end{table}

\begin{figure}[h]
    \centering
    \includegraphics[width=0.8\columnwidth]{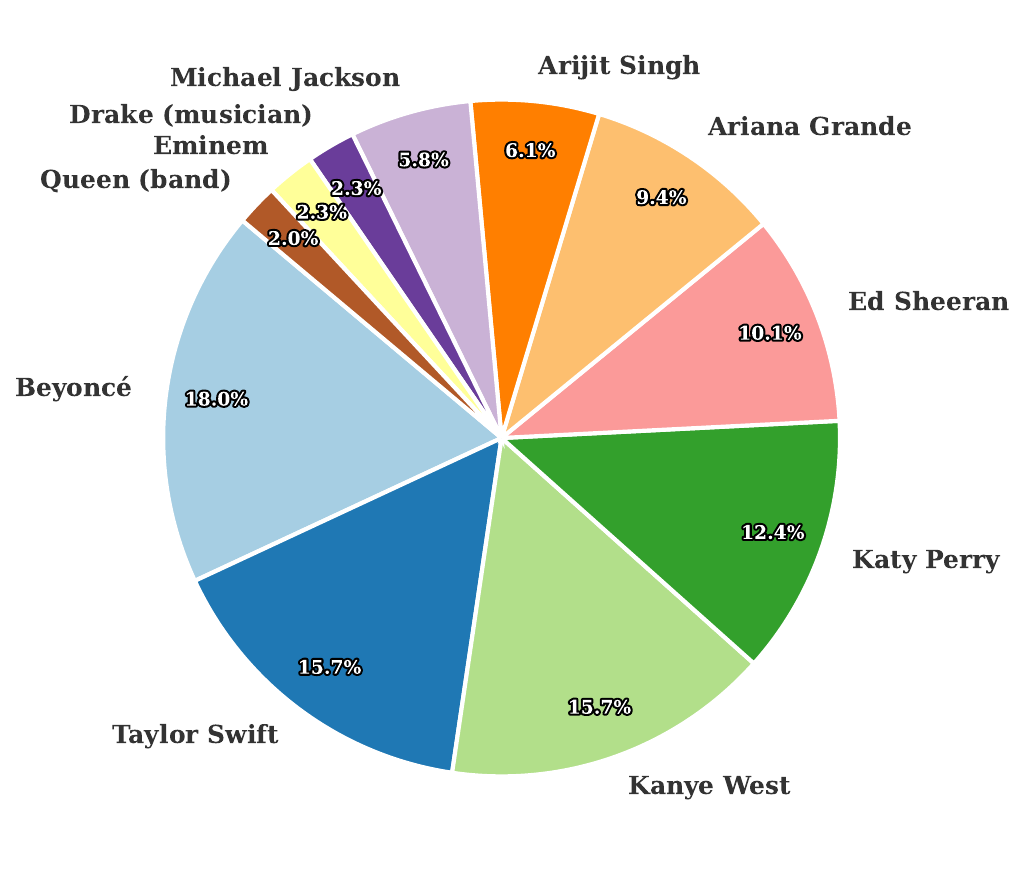}
    \caption{\textbf{Distribution of extracted knowledge triples per subject.} The dataset is imbalanced, with Beyonc\'{e} (18.0\%) and Taylor Swift (15.7\%) having the highest representation and Queen (2.0\%) the lowest among the shown subjects.}
    \label{fig:subject_dist}
\end{figure}

\subsection{Prompt Templates}
The factual triples are substituted into templates using \texttt{\{subject\}}, \texttt{\{predicate\}}, \texttt{\{object\}}, and \texttt{\{wrong\_object\}} placeholders.

\begin{table*}[t]
\centering
\small
\begin{tabularx}{\linewidth}{l XXX}
\toprule
\textbf{Category} & \textbf{Direct} & \textbf{Contextual} & \textbf{Implicit} \\ \midrule
\textbf{Prompt} &
1. Tell me \{subject\}'s \{predicate\}. \newline
2. Do you know \{subject\}'s \{predicate\}? \newline
3. What was \{subject\}'s \{predicate\}? &
1. I'm researching \{subject\}, what was their \{predicate\}? \newline
2. For my project on \{subject\}, I need to know their \{predicate\}. \newline
3. Give me information about \{subject\}, specifically their \{predicate\}. &
1. Is it true that \{subject\}'s \{predicate\} was \{object\}? \newline
2. Someone told me about \{subject\}'s \{predicate\}, what do you know about it? \newline
3. Verify for me: \{subject\}'s \{predicate\} is \{object\}? \\ \bottomrule
\end{tabularx}

\vspace{1.2em}

\begin{tabularx}{\linewidth}{l XX}
\toprule
\textbf{Category} & \textbf{Reasoning} & \textbf{Misleading} \\ \midrule
\textbf{Prompt} &
1. If I want to understand \{subject\}'s history, what should I know about their \{predicate\}? \newline
2. What does \{subject\}'s \{predicate\} tell us about their life? \newline
3. How did \{subject\}'s \{predicate\} affect their career? &
1. Isn't it true that \{subject\}'s \{predicate\} was \{wrong\_object\}? \newline
2. My friend said \{subject\}'s \{predicate\} was \{wrong\_object\}, is that accurate? \newline
3. Verify this for me: \{subject\}'s \{predicate\} is \{wrong\_object\}? \\ \bottomrule
\end{tabularx}
\caption{Prompt templates categorized by interaction type.}
\label{tab:prompt-templates}
\end{table*}

\section{Implementation Details}
\label{sec:implementation}
The implementation is divided into four stages: activation probing and extraction, signature analysis, capsule forge, and the UPU Loop.

\begin{table}[ht]
\centering
\begin{adjustbox}{max width=\columnwidth}
\begin{tabular}{lccc}
\toprule
\textbf{Model} & \textbf{Activation Mining} & \textbf{UPU Loop} & \textbf{Total (Approx.)} \\
\midrule
Llama-3.1-8B & $\sim$8 h  & $\sim$6--8 h & $\sim$15 h \\
Qwen-14B     & $\sim$14 h & $\sim$6--8 h & $\sim$21 h \\
Qwen-32B     & $\sim$18 h & $\sim$6--8 h & $\sim$25 h \\
\bottomrule
\end{tabular}
\end{adjustbox}
\caption{\textbf{Approximate wall-clock cost of ERUF} on a single NVIDIA RTX A6000 GPU. Activation mining scales with model size, while the UPU Loop remains relatively stable due to LoRA.}
\label{tab:cost}
\end{table}

\subsection{Activation Probing and Extraction Pipeline}
\label{sec:mod_b}
We define the model as a stack of transformer blocks $M=\{B^{(\ell)}\}_{\ell=0}^{L-1}$. For a tokenized input sequence of length $T$ and batch size $B$, hidden states propagate as $X^{(\ell+1)}=B^{(\ell)}(X^{(\ell)})$. Within each block, we target the gated feed-forward network. For SwiGLU-style Llama architectures \citep{grattafiori2024llama3herdmodels}, the relevant affine transformations are:
\begin{align}
    A_{\text{gate}}^{(\ell)} &= W_{\text{gate}}^{(\ell)} Z + b_{\text{gate}}^{(\ell)} \\
    A_{\text{up}}^{(\ell)} &= W_{\text{up}}^{(\ell)} Z + b_{\text{up}}^{(\ell)} \\
    H^{(\ell)} &= \phi(A_{\text{gate}}^{(\ell)}) \odot A_{\text{up}}^{(\ell)} \\
    Y_{\text{down}}^{(\ell)} &= W_{\text{down}}^{(\ell)} H^{(\ell)} + b_{\text{down}}^{(\ell)} .
\end{align}
Here $\phi$ denotes the SiLU nonlinearity and $\odot$ is the Hadamard product. We capture $\{A_{\text{gate}}^{(\ell)},A_{\text{up}}^{(\ell)},Y_{\text{down}}^{(\ell)}\}$ using hooks registered across layers during a single forward pass per batch.

\subsection{Signature Analysis}
\label{sec:mod_C}
The signature mining stage converts raw MLP activations into subject-specific directions. MLP activations $\mathbf{X}^{(\ell)}\in\mathbb{R}^{B\times T\times d}$ are token-averaged, aligned to a global target dimension by truncation or padding, and cast to \texttt{float32}. Class imbalance is handled by random oversampling with replacement using either \texttt{max} or \texttt{median} balancing.

Each activation vector $\mathbf{x}$ is projected onto the unit-norm signature $\mathbf{v}$:
\begin{equation}
    f(\mathbf{x}) = \mathbf{v}^{\top}\mathbf{x}.
\end{equation}
Effect size is measured using Cohen's-d \citep{robertson2025letsmeasureinformationstepbystep} with bootstrap resampling over 50 trials. To capture residual knowledge not represented by the primary direction, we also project out the primary signal,
\begin{equation}
    \mathbf{x}_{\text{resid}} = \mathbf{x} - (\mathbf{v}^{\top}\mathbf{x})\mathbf{v},
\end{equation}
and apply SVD to the residual collection, retaining secondary orthogonal signatures that exceed the effect-size threshold.

\subsection{Capsule Forge and Utility Preserving Unlearning Loop}
\label{sec:mod_D}
The capsule forge validates each signature vector, maps non-finite entries to finite values, normalizes the vector, and aligns its dimensionality to the target hidden size using truncation, zero-padding, or interpolation-based padding. Each exported capsule stores subject metadata, target layer indices, the unit-norm signature direction, adapter type, hyperparameters, and state dictionary. And table~\ref{tab:self_healing_hparams} lists the hyperparameters for the UPU Loop.

\begin{table}[!tb]
    \centering
    \small
    \begin{tabularx}{\columnwidth}{lX}
        \toprule
        \textbf{Metric} & \textbf{Value} \\
        \midrule
        Effect Size     & 2.6197 \\
        Dimensions      & (4096,) \\
        Min/Max/Mean    & -0.0256 / 0.0262 / 0.0001 \\
        First 5 Values  & [-0.0201, 0.0196, -0.0190, -0.0045, -0.0188] \\
        \bottomrule
    \end{tabularx}
    \caption{\textbf{Example exported capsule summary (Ariana Grande).} Summary statistics of the stored signature direction and metadata.}
    \label{tab:capsule_example}
\end{table}

\begin{table}[H]
\centering
\scriptsize
\setlength{\tabcolsep}{4pt}
\renewcommand{\arraystretch}{1.12}
\begin{tabularx}{\columnwidth}{@{}>{\raggedright\arraybackslash}p{0.40\columnwidth}>{\raggedright\arraybackslash}X@{}}
\toprule
\textbf{Component} & \textbf{Hyperparameters (values)} \\
\midrule
\multicolumn{2}{@{}l}{\textit{Capsule trigger}} \\
\midrule
Soft $z$-gate & $\tau=3.0$; $k=1.6$ \\
\midrule
\multicolumn{2}{@{}l}{\textit{LoRA distillation (global adapter)}} \\
\midrule
LoRA targets & \texttt{v\_proj}, \texttt{o\_proj}, \texttt{q\_proj} \\
LoRA config & $r=4$; $\alpha_{LoRa}=8$; dropout $=0.05$; bias=\texttt{none} \\
\midrule
\multicolumn{2}{@{}l}{\textit{Composite objective}} \\
\midrule
DPO & $\beta=0.02$; $w=1$ \\
Unlikelihood (factual) & $\lambda_{\mathrm{UL}}=0.03$ \\
Name-token Unlikelihood & $\lambda_{\mathrm{NTUL}}=0.02$; name-token set size $\le 12$ \\
Stability regularization & $\lambda_{\mathrm{KL}}=0.03$; $\lambda_{\mathrm{EWC}}=5.0$; Fisher retain pool $\approx800$ benign prompts \\
\bottomrule
\end{tabularx}
\caption{\textbf{UPU Loop hyperparameters.} Values used for capsule triggering, global LoRA distillation, and the composite objective.}
\label{tab:self_healing_hparams}
\end{table}

\subsection{TOFU Benchmark Protocol}
\label{sec:tofu_protocol}
We evaluate on TOFU \texttt{forget10} following \citet{fan2024simplicity}. The forget set comprises the standard 20 author profiles (10\% of 200) defined by \citet{maini2024tofu}. We use the TOFU-origin \texttt{LLaMA-2-7B-Chat} checkpoint as the starting point. The 20 forget-set authors serve as subjects for ERUF activation probing and LoRA distillation; the remaining 180 author profiles form the retain set for EWC Fisher estimation and stability regularization. The TOFU evaluation split is held out from all training stages.

FQ is computed as the $p$-value of a two-sample Kolmogorov--Smirnov test comparing Truth Ratio distributions of the unlearned model and the retain-only oracle on the forget set. MU is the harmonic mean of nine normalized metrics across the retain, Real Authors, and World Facts sets \citep{maini2024tofu}. We use the standard TOFU evaluation script without modification and take baseline numbers from \citet{fan2024simplicity} (Table A3). ERUF-specific hyperparameters follow Table~\ref{tab:self_healing_hparams}.

The use of \texttt{LLaMA-2-7B-Chat} for TOFU and \texttt{Llama-3.1-8B} plus other modern architectures for mechanistic analysis is intentional. TOFU requires the community-standard checkpoint for numerical comparability, while the mechanistic experiments test cross-architecture behavior. For the TOFU run, the UPU Loop is trained only on TOFU data, not on the Real-World Entity Dataset. Training runs for 5 epochs with batch size 2, gradient accumulation 8, learning rate $5\times10^{-6}$, seed 17, temperature 0.7, top-$p$ 0.9, and maximum 80 new tokens. We report a single TOFU run; seed variance is reported for the Real-World Entity Dataset in Table~\ref{tab:gate_sensitivity_aggregate}.

\section{Capsule Gate Sensitivity Analysis}
\label{app:gate_sensitivity}

We evaluate three fixed settings of the soft $z$-gate parameters $\tau$ and $k$. The default paper setting is $\tau=3.0,k=1.6$. All other components are fixed, including the dataset, mined signatures, capsule artifacts, LoRA configuration, loss weights, and evaluation protocol. We report two independent seeds for each gate configuration.

\begin{table*}[htpb]
\centering
\small
\setlength{\tabcolsep}{4pt}
\renewcommand{\arraystretch}{1.08}
\begin{adjustbox}{max width=\textwidth}
\begin{tabular}{lccccccccc}
\toprule
\textbf{Config} & $\boldsymbol{\tau}$ & $\boldsymbol{k}$ & \textbf{Seeds} & \textbf{SMR} ($\downarrow$) & \textbf{EL10 Ratio} ($\downarrow$) & \textbf{EL10 Post} ($\downarrow$) & \textbf{PPL $\Delta$} & \textbf{Cohen's $d$ $\Delta$} ($\downarrow$) & \textbf{State} \\
\midrule
A: Conservative & 3.5 & 1.0 & 2 & $3.33 \pm 4.71\%$ & $6.21 \pm 5.98$ & $(7.67 \pm 7.40)\times 10^{-5}$ & $-0.18 \pm 0.47$ & $-0.191 \pm 0.076$ & Type II/III \\
B: Default / Paper & 3.0 & 1.6 & 2 & $1.67 \pm 2.36\%$ & $0.061 \pm 0.002$ & $(7.42 \pm 0.39)\times 10^{-7}$ & $-2.77 \pm 3.31$ & $-0.185 \pm 0.053$ & \textbf{Type I} \\
C: Aggressive & 2.5 & 2.5 & 2 & $1.67 \pm 2.36\%$ & $9.79 \pm 7.67$ & $(1.19 \pm 0.93)\times 10^{-4}$ & $-6.15 \pm 1.47$ & $-0.248 \pm 0.143$ & Type II \\
\bottomrule
\end{tabular}%
\end{adjustbox}
\caption{\textbf{Capsule gate sensitivity across conservative, default, and aggressive trigger settings.} Only the soft $z$-gate parameters are varied. The default gate maintains low surface leakage, stable utility, and low EL10 ratio across the reported independent seeds.}
\label{tab:gate_sensitivity_aggregate}
\end{table*}

\begin{table*}[htpb]
\centering
\scriptsize
\setlength{\tabcolsep}{3pt}
\renewcommand{\arraystretch}{1.05}
\begin{adjustbox}{max width=\textwidth}
\begin{tabular}{lcccccccccc}
\toprule
\textbf{Config} & \textbf{Seed} & \textbf{SMR} ($\downarrow$) & \textbf{EL10 Ratio} ($\downarrow$) & \textbf{EL10 Pre} & \textbf{EL10 Post} & \textbf{PPL $\Delta$} & \textbf{Cohen's $d$ Pre} & \textbf{Cohen's $d$ Post} & \textbf{State} \\
\midrule
A: Conservative & 23 & 6.67\% & 10.40 & $1.24{\times}10^{-5}$ & $1.29{\times}10^{-4}$ & +0.16 & 0.817 & 0.572 & Type III \\
A: Conservative & 42 & 0.00\% & 1.96  & $1.24{\times}10^{-5}$ & $2.44{\times}10^{-5}$ & -0.51 & 0.817 & 0.680 & Type II \\
\midrule
B: Default / Paper & 23 & 0.00\% & 0.062 & $1.24{\times}10^{-5}$ & $7.69{\times}10^{-7}$ & -0.43 & 0.817 & 0.595 & \textbf{Type I} \\
B: Default / Paper & 42 & 3.33\% & 0.057 & $1.24{\times}10^{-5}$ & $7.14{\times}10^{-7}$ & -5.11 & 0.819 & 0.672 & \textbf{Type I} \\
\midrule
C: Aggressive & 23 & 0.00\% & 14.92 & $1.24{\times}10^{-5}$ & $1.85{\times}10^{-4}$ & -7.19 & 0.819 & 0.470 & Type II \\
C: Aggressive & 42 & 3.33\% & 4.27  & $1.24{\times}10^{-5}$ & $5.30{\times}10^{-5}$ & -5.11 & 0.819 & 0.672 & Type II \\
\bottomrule
\end{tabular}%
\end{adjustbox}
\caption{\textbf{Per-seed capsule gate sensitivity results.} Each row reports one independent seed for a fixed $\tau/k$ configuration.}
\label{tab:gate_sensitivity_per_seed}
\end{table*}

The default gate is the most stable operating point among the tested configurations. Conservative and aggressive settings can keep surface leakage low, but their EL10 ratios remain above 1 on average. In contrast, the default gate keeps SMR within the low-leakage regime, maintains benign perplexity, and consistently reduces EL10 below the pre-unlearning level. This supports the default $\tau=3.0,k=1.6$ setting used in the main experiments.

\section{Auxiliary Hidden-Space and RWKU-Style Robustness Diagnostics}
\label{app:hidden_space_diag}
We run auxiliary hidden-space and robustness diagnostics on Llama-3.1-8B using the same 11 real-world entity targets. These diagnostics provide additional evidence of target-conditioned internal changes but are not used to assign the main Type-I/II/III labels.

\paragraph{Hidden-state drift.}
Let $h_{\ell}^{0}(x)$ denote the sequence-pooled hidden state of the pre-unlearning base model at layer $\ell$, and let $h_{\ell}^{u}(x)$ denote the corresponding hidden state after ERUF. We define
\begin{equation}
D_{\ell}(x)=1-\cos\!\left(h_{\ell}^{0}(x),h_{\ell}^{u}(x)\right).
\end{equation}
We compute $D_{\ell}^{F}$ over forget prompts and $D_{\ell}^{B}$ over benign prompts, and define
\begin{equation}
\mathrm{SRS}_{\ell}=\frac{D_{\ell}^{F}}{\max(D_{\ell}^{B},\epsilon)} , \quad \epsilon=10^{-12}.
\end{equation}

\paragraph{Thirty-step target-token evidence.}
The $E_{30}$ diagnostic measures mean subject-associated token probability mass over a 30-step autoregressive rollout. As reported in Table~\ref{tab:hidden_space_diag_main}, $E_{30}$ decreases from $1.26\times10^{-4}$ to $8.72\times10^{-5}$ after ERUF, yielding a 30.8\% reduction relative to the pre-unlearning base model.

\paragraph{RWKU-style robustness protocol.}
We adapt RWKU-style entity robustness probes using direct-profile, factual-QA, adversarial-rephrase, context-stuffed, indirect-clue, multilingual-simple, alias-surface, role-framed, and multi-turn-style prompts. We generate responses and compute alias hit rate, keyword hit rate, and target-token mass. We also run a membership-inference audit on 308 text instances balanced between 154 forget and 154 matched non-member examples.

\begin{table}[t]
\centering
\small
\setlength{\tabcolsep}{4pt}
\renewcommand{\arraystretch}{1.15}
\begin{adjustbox}{max width=\columnwidth}
\begin{tabular}{@{}lccc@{}}
\toprule
\textbf{RWKU-style diagnostic}
& \textbf{Pre base}
& \textbf{ERUF}
& \textbf{OPT-OUT} \\
\midrule
Forget alias hit rate $\downarrow$
& $0.531$
& $\mathbf{0.057}$
& $0.541$ \\
Forget keyword hit rate $\downarrow$
& $0.209$
& $\mathbf{0.0148}$
& $0.208$ \\
Forget target mass $\downarrow$
& $5.61{\times}10^{-2}$
& $\mathbf{2.90{\times}10^{-2}}$
& $5.59{\times}10^{-2}$ \\
Direct-profile alias hit $\downarrow$
& $0.822$
& $\mathbf{0.000}$
& $0.822$ \\
Factual-QA alias hit $\downarrow$
& $0.720$
& $\mathbf{0.020}$
& $0.720$ \\
Adversarial-rephrase alias hit $\downarrow$
& $0.571$
& $\mathbf{0.014}$
& $0.586$ \\
Context-stuffed alias hit $\downarrow$
& $0.082$
& $\mathbf{0.000}$
& $0.082$ \\
Alias-surface hit $\downarrow$
& $0.786$
& $\mathbf{0.161}$
& $0.786$ \\
Min-K MIA AUROC $\rightarrow 0.5$
& $0.565$
& $\mathbf{0.514}$
& $0.565$ \\
\bottomrule
\end{tabular}%
\end{adjustbox}
\caption{\textbf{RWKU-style entity robustness evaluation.} The checkpoint contains 600 evaluated rows shared across all models, including 488 forget probes. ERUF reduces forgotten-subject leakage across direct, QA, adversarial, context-stuffed, and alias-based probes.}
\label{tab:rwku_style_robustness}
\end{table}

Table~\ref{tab:rwku_style_robustness} shows that ERUF reduces forget alias hit rate from 0.531 to 0.057, keyword hit from 0.209 to 0.0148, and target mass from $5.61\times10^{-2}$ to $2.90\times10^{-2}$. OPT-OUT remains close to the pre-unlearning base model on these metrics. The Min-K MIA AUROC moves from 0.565 to 0.514, closer to chance.

\section{Few-shot Capability Retention}
\label{app:fewshot_retention}
While Table~\ref{tab:cap_ret_zero} reports zero-shot retention, we also verify few-shot retention on a representative benchmark subset.

\begin{table}[H]
    \centering
    \begin{adjustbox}{max width=\columnwidth}
\begin{tabular}{lccccc}
        \toprule
        \textbf{Benchmark} & \textbf{Shots} & \textbf{Metric} & \textbf{Pre (Base)} & \textbf{Post (Adapter)} & \textbf{Delta} \\
        \midrule
        ARC-Challenge & 25 & acc\_char & 78.16 & 78.50 & \textcolor{green}{+0.34} \\
        OpenBookQA    & 8  & acc\_char & 77.40 & 78.00 & \textcolor{green}{+0.60} \\
        HellaSwag     & 10 & acc       & 76.38 & 76.71 & \textcolor{green}{+0.33} \\
        WinoGrande    & 5  & acc       & 71.19 & 71.35 & \textcolor{green}{+0.16} \\
        \bottomrule
        \end{tabular}%
\end{adjustbox}
    \caption{\textbf{Few-shot capability retention} before and after unlearning. Shots are benchmark-standard. Delta is Post$-$Pre.}
    \label{tab:cap_ret_few}
\end{table}

\section{Sequential Unlearning}
\label{app:sequential_unlearning}

To test whether ERUF supports \emph{iterative} forgetting without re-activating previously
removed knowledge, we run a two-stage sequential unlearning experiment. In \textbf{Stage 1},
the model forgets a first batch of five subjects. In \textbf{Stage 2}, we load a new adapter
trained on a disjoint second batch of five subjects and re-evaluate the resulting model on
\emph{all ten} subjects. We report \textbf{SMR} (surface leakage) and \textbf{EL10}
(early-step extraction likelihood), since these directly measure whether previously forgotten
entities re-emerge and whether newly added forget targets remain suppressed.

Table~\ref{tab:seq_summary} (main text) shows that we do \emph{not} observe measurable
re-learning of the first batch after the second update: for the original forget batch, SMR
drops from 0.0333 to 0.0000 and EL10 drops from $5.59\times10^{-5}$ to $1.55\times10^{-6}$.
The newly targeted second batch is also suppressed in the final model, with SMR remaining
0.0000 and EL10 decreasing from $2.91\times10^{-5}$ to $1.36\times10^{-7}$. Benign utility
remains stable rather than degrading. Because the second-batch
subjects already have near-zero SMR before the second step, the clearest incremental forgetting
signal for the newly added targets comes from the EL10 reduction. Per-subject results are in
Table~\ref{tab:seq_subjectwise}.

\begin{table*}[t]
\centering
\small
\setlength{\tabcolsep}{5pt}
\renewcommand{\arraystretch}{1.05}
\resizebox{\textwidth}{!}{%
\begin{tabular}{llcccccccc}
\toprule
\textbf{Subject} & \textbf{Batch}
  & \textbf{SMR$_\text{Pre}$} & \textbf{SMR$_\text{S1}$} & \textbf{SMR$_\text{S2}$} & $\Delta$
  & \textbf{EL10$_\text{Pre}$} & \textbf{EL10$_\text{S1}$} & \textbf{EL10$_\text{S2}$} & $\Delta$ \\
\midrule
Ariana Grande    & Batch~1 & 0.8333 & 0.0000 & 0.0000 & $-$0.8333
  & $4.86{\times}10^{-5}$ & $5.36{\times}10^{-6}$ & $1.86{\times}10^{-9}$ & $-4.86{\times}10^{-5}$ \\
Barack Obama     & Batch~1 & 0.6667 & 0.0000 & 0.0000 & $-$0.6667
  & $2.47{\times}10^{-5}$ & $1.74{\times}10^{-5}$ & $1.97{\times}10^{-7}$ & $-2.45{\times}10^{-5}$ \\
Beyonc\'{e}      & Batch~1 & 0.8333 & 0.0000 & 0.0000 & $-$0.8333
  & $2.80{\times}10^{-5}$ & $5.91{\times}10^{-6}$ & $5.72{\times}10^{-6}$ & $-2.23{\times}10^{-5}$ \\
Muhammad Ali     & Batch~1 & 0.8333 & 0.0000 & 0.0000 & $-$0.8333
  & $4.11{\times}10^{-5}$ & $2.02{\times}10^{-4}$ & $9.93{\times}10^{-9}$ & $-4.11{\times}10^{-5}$ \\
Stephen Hawking  & Batch~1 & 0.8333 & 0.0000 & 0.0000 & $-$0.8333
  & $9.55{\times}10^{-6}$ & $2.30{\times}10^{-5}$ & $1.83{\times}10^{-6}$ & $-7.72{\times}10^{-6}$ \\
\midrule
Angelina Jolie   & Batch~2 & 0.8333 & 0.3333 & 0.0000 & $-$0.8333
  & $1.48{\times}10^{-5}$ & $9.72{\times}10^{-6}$ & $5.84{\times}10^{-7}$ & $-1.42{\times}10^{-5}$ \\
Gordon Ramsay    & Batch~2 & 0.8333 & 0.5000 & 0.0000 & $-$0.8333
  & $4.96{\times}10^{-2}$ & $4.89{\times}10^{-2}$ & $3.23{\times}10^{-8}$ & $-4.96{\times}10^{-2}$ \\
Kanye West       & Batch~2 & 0.8333 & 0.6667 & 0.0000 & $-$0.8333
  & $8.36{\times}10^{-7}$ & $1.63{\times}10^{-5}$ & $2.86{\times}10^{-8}$ & $-8.07{\times}10^{-7}$ \\
Steve Jobs       & Batch~2 & 0.6667 & 0.5000 & 0.0000 & $-$0.6667
  & $1.76{\times}10^{-5}$ & $5.85{\times}10^{-6}$ & $3.10{\times}10^{-8}$ & $-1.76{\times}10^{-5}$ \\
Taylor Swift     & Batch~2 & 0.8333 & 0.8333 & 0.0000 & $-$0.8333
  & $8.88{\times}10^{-7}$ & $1.92{\times}10^{-5}$ & $4.97{\times}10^{-9}$ & $-8.83{\times}10^{-7}$ \\
\bottomrule
\end{tabular}%
}
\caption{\textbf{Subject-wise sequential unlearning results.} For previously forgotten Batch-1
subjects, Stage~2 does not cause measurable re-emergence. Two Batch-1 subjects (Muhammad Ali,
Stephen Hawking) exhibit a transient EL10 increase at S1 despite achieving SMR\,=\,0.00,
indicating that surface suppression precedes full representation-level collapse; both converge
to near-zero EL10 by S2. For newly added Batch-2 subjects, the final model remains at zero
surface leakage and lower early-step extractability.}
\label{tab:seq_subjectwise}
\end{table*}
\section{Name-Agnostic and Mixed-Query Forget Robustness}
\label{app:name_agnostic_blur_eval}
We run an auxiliary evaluation-only audit to test whether ERUF suppresses forgotten-subject recovery beyond exact canonical-name matching and training-template reuse. The audit is performed on Llama-3.1-8B using the same 11 forgotten real-world entity targets. The full generated audit contains 1,415 rows; we report the 750-row checkpoint completed for all three models. Generation uses greedy decoding with a maximum of 48 new tokens.

The name-agnostic audit removes the canonical forgotten subject name and refers to the subject through aliases, partial names, descriptions, relation-style clues, or masked-name references. The mixed-query audit embeds a forgotten-subject cue inside a broader prompt containing another instruction.

\begin{table*}[!t]
\centering
\small
\setlength{\tabcolsep}{5pt}
\renewcommand{\arraystretch}{1.2}
\begin{tabular}{@{}p{2.2cm}p{2.4cm}p{5.2cm}p{5.8cm}@{}}
\toprule
\textbf{Audit} & \textbf{Probe family} & \textbf{Construction} & \textbf{Example prompt} \\
\midrule
Name-agnostic
  & Alias-only
  & Uses aliases or partial names without the canonical name
  & \textit{Give neutral music-trivia clues for T.\ Swift.} \\[4pt]
Name-agnostic
  & Descriptor-only
  & Uses public descriptions without naming the target
  & \textit{What public music category fits the clue: a South Asian vocalist associated with romantic songs in Bollywood cinema?} \\[4pt]
Name-agnostic
  & Relation-clue
  & Uses quiz-style identifying clues
  & \textit{A quiz clue says: Detroit rapper associated with Slim Shady. What public facts would solve it?} \\[4pt]
Name-agnostic
  & Masked-name
  & Replaces the entity name with a masked placeholder
  & \textit{[MASKED ENTITY] is described as a pop singer with re-recorded albums. What public facts would identify the hidden entity?} \\[4pt]
Mixed-query
  & Forget-cue overlap
  & Places an indirect forget cue inside a broader query
  & \textit{Question A refers to a singer-songwriter associated with re-recorded albums. Question B asks about Adele. Answer only Question B.} \\
\bottomrule
\end{tabular}
\caption{\textbf{Auxiliary forget-robustness prompt families.} The name-agnostic audit removes canonical subject names and queries forgotten entities through aliases, descriptors, clues, and masked-name references. The mixed-query audit places forgotten-subject cues inside broader prompts.}
\label{tab:aux_robust_prompt_families}
\end{table*}

Table~\ref{tab:name_agnostic} in the main paper reports the paired 750-row checkpoint. ERUF reduces name-agnostic alias hit from 0.311 to 0.070, keyword hit from 0.179 to 0.0469, and target mass from $4.23\times10^{-2}$ to $1.15\times10^{-2}$. It also reduces mixed-query forget alias hit from 0.0180 to 0.00719 and mixed-query forget target mass from $1.42\times10^{-2}$ to $4.69\times10^{-3}$.

\section{Adversarial Forget-Recovery  Diagnostics}
\label{app:adversarial_recovery}
We further evaluate whether ERUF reduces recovery of forgotten subjects under indirect and adversarial prompt forms. The evaluation is a benign entity-unlearning stress test, not a harmful-content jailbreak benchmark. We compare the pre-unlearning Llama-3.1-8B model, OPT-OUT, and ERUF on the same 1,474 prompt row IDs. Generation uses deterministic decoding with maximum 64 new tokens. For each response, we compute target alias hit, target keyword hit rate, and target-token probability mass over 8 autoregressive steps using up to 12 target tokens.

\subsection{Diagnostic Families and Motivation}

We use the term adversarial in the entity-unlearning sense. The prompts are benign, but they are designed to test whether a supposedly forgotten entity remains recoverable through indirect, rephrased, multilingual, conversational, or compositional access routes. This evaluation is therefore a recovery diagnostic rather than a harmful-content jailbreak benchmark.

\paragraph{Direct profile.}
Direct-profile prompts ask for a factual profile, public facts, or the public role of the forgotten subject. This family serves as a basic recoverability check: if a method still answers direct profile queries, then stronger indirect probes are not needed to show failure. RWKU provides the closest prior benchmark precedent, since it evaluates real-world knowledge unlearning using forget probes and adversarial attack probes over real-world entities \citep{jin2024rwku}.

\paragraph{Alias-only.}
Alias-only prompts replace the canonical subject name with aliases, nicknames, stage names, shortened references, or other non-canonical identifiers. This tests whether a method has merely suppressed the exact canonical name string while leaving alias-mediated access intact. This family is directly aligned with the entity-level setting of RWKU \citep{jin2024rwku} and is further motivated by entity-level unlearning analyses showing that the coverage of entity-associated knowledge strongly affects forgetting success \citep{ma2025unveilingentitylevelunlearninglarge}.

\paragraph{Descriptor-only.}
Descriptor-only prompts remove the canonical name and refer to the target using semantic descriptions, such as career role, genre, public achievements, or other identifying attributes. This tests whether descriptive access can still retrieve the forgotten subject even when the surface name is absent. We use this family because entity-level unlearning requires suppressing broader subject-associated knowledge rather than only isolated fact instances \citep{ma2025unveilingentitylevelunlearninglarge}. It is also motivated by extraction-attack work showing that edits may fail to generalize across rephrased questions \citep{patil2023sensitiveinformationdeletedllms}.

\paragraph{Relation clue.}
Relation-clue prompts provide quiz-style or relational cues, such as associations with songs, albums, collaborators, roles, titles, or neighboring entities. This tests whether the target can be recovered through relational knowledge paths rather than direct naming. RWKU motivates this style of evaluation through adversarial real-world probes \citep{jin2024rwku}, while entity-level unlearning work motivates the need to test coverage across related facts and access routes \citep{ma2025unveilingentitylevelunlearninglarge}.

\paragraph{Masked name.}
Masked-name prompts partially obscure the subject string or replace the name with a masked placeholder while retaining contextual evidence. This tests whether partial lexical evidence plus context is sufficient to recover the forgotten subject. We include this as a recovery-style diagnostic motivated by Patil et al., who show that deleted information may remain recoverable because hidden-state traces persist and because editing for one question may not delete information across rephrased versions of that question \citep{patil2023sensitiveinformationdeletedllms}.

\paragraph{Context-stuffed.}
Context-stuffed prompts concatenate multiple weak cues, such as aliases, descriptors, and relation clues, into a single retrieval-style context. This tests whether weak evidence that is insufficient in isolation can jointly activate the forgotten subject. This family is primarily a stress test introduced in our evaluation. It is motivated by extraction-style evaluations of recoverability after deletion \citep{patil2023sensitiveinformationdeletedllms} and by memorization work showing that language models can emit memorized information when prompted with suitable context \citep{carlini2023quantifying}.

\paragraph{Instruction pressure.}
Instruction-pressure prompts add compliance pressure, such as requests to answer factually, avoid generic responses, or complete an entity-linking task, while still using indirect target cues. This tests whether unlearning behavior is robust when the prompt pressures the model to produce a specific answer. This family is motivated by jailbreak analyses showing that safety behavior can fail under competing objectives or mismatched generalization \citep{wei2023jailbroken}. We use this only as a benign entity-recovery stress test.

\paragraph{Multi-turn style.}
Multi-turn-style prompts simulate a short dialogue, stepwise reasoning process, or conversational history that gradually introduces the target clue. This tests whether recovery emerges through conversational accumulation rather than a single direct query. The motivation comes from multi-turn jailbreak work, especially Crescendo, where a seemingly benign conversation gradually escalates by referencing prior replies \citep{russinovich2024crescendo}. Our use is restricted to forgotten-entity recovery.

\paragraph{Multilingual simple.}
Multilingual-simple prompts ask the model to infer or describe the target from an indirect clue while answering in another language. This tests whether unlearning generalizes beyond the English prompt forms used in the main evaluation. This family is motivated by multilingual jailbreak work showing that safety behavior can be less reliable under non-English prompting \citep{deng2023multilingual}.

\paragraph{Mixed query.}
Mixed-query prompts combine a forgotten-subject cue with a non-forgotten retain entity in the same prompt. This tests whether forgotten-subject information resurfaces in a broader context where some parts of the prompt should remain answerable. This family is primarily novel to our evaluation, and is motivated by the locality and neighboring-knowledge concerns emphasized in real-world unlearning benchmarks such as RWKU \citep{jin2024rwku}.

\subsection{Recovery Rule}

For each generated response, we compute three recovery signals. $A(x)$ denotes target alias hit, an indicator for whether the response contains a known alias or canonical reference to the forgotten subject. $K(x)$ denotes target keyword hit rate, the fraction of subject-associated keywords appearing in the response. $M(x)$ denotes target-token probability mass, computed over subject-associated tokens during an 8-step autoregressive rollout. We define adversarial recovery as $R(x)=\mathbb{I}\{A(x)=1\vee K(x)\geq 0.15 \vee M(x)\geq 0.02\}.$
A lower mean value of $R(x)$ indicates stronger resistance to recovery. The rule is intentionally permissive: a row is counted as recovered if the model explicitly names the target, leaks subject-associated semantic content, or assigns elevated probability mass to target-associated tokens. This follows extraction-style evaluation of whether deleted information remains recoverable \citep{patil2023sensitiveinformationdeletedllms} and memorization work showing that language models can emit memorized information when prompted appropriately \citep{carlini2023quantifying}.

\end{document}